\documentclass[runningheads]{llncs}
\usepackage[T1]{fontenc}
\usepackage{amsmath}
\usepackage{subcaption}
\usepackage{amsfonts}\usepackage[section]{placeins}
\usepackage{float}
\usepackage{orcidlink}
\usepackage{graphicx}
\usepackage{color}
\begin{document}
%
\title{Mechanistic Interpretability for
Transformer-based Time Series Classification}

\titlerunning{Mechanistic Interpretability for Time Series Transformers}
%
%

\author{
Matīss Kalnāre\orcidlink{0009-0003-8610-7481} \and
Sofoklis Kitharidis\orcidlink{0009-0005-8404-0724} \and
Thomas B{\"a}ck\orcidlink{0000-0001-6768-1478} \and
Niki van Stein\orcidlink{0000-0002-0013-7969}
}

\authorrunning{Kalnāre et al.}

\institute{
Leiden Institute of Advanced Computer Science (LIACS), \\
Leiden University, Einsteinweg 55, \\
2333 CC Leiden, The Netherlands \\
\email{\{m.kalnare, s.kitharidis, T.H.W.Baeck, n.van.stein\}@liacs.leidenuniv.nl}
}
\maketitle              
\begin{abstract}
Transformer-based models have become state-of-the-art tools in various machine learning tasks, including time series classification, yet their complexity makes understanding their internal decision-making challenging. Existing explainability methods often focus on input-output attributions, leaving the internal mechanisms largely opaque. This paper addresses this gap by adapting various Mechanistic Interpretability techniques; activation patching, attention saliency, and sparse autoencoders, from NLP to transformer architectures designed explicitly for time series classification. We systematically probe the internal causal roles of individual attention heads and timesteps, revealing causal structures within these models. Through experimentation on a benchmark time series dataset, we construct causal graphs illustrating how information propagates internally, highlighting key attention heads and temporal positions driving correct classifications. Additionally, we demonstrate the potential of sparse autoencoders for uncovering interpretable latent features. Our findings provide both methodological contributions to transformer interpretability and novel insights into the functional mechanics underlying transformer performance in time series classification tasks.


\end{abstract}

\section{Introduction}
\label{sec:introduction}

Deep learning models, ranging from convolutional networks to modern Transformer architectures, have achieved remarkable accuracy in domains as diverse as computer vision, natural language processing, and time series analysis. Yet this performance often comes at the cost of interpretability: practitioners and stakeholders lack insight into the internal decision‐making processes of these “black‐box” models, undermining trust in safety‐critical applications such as healthcare, finance, and autonomous systems.

Explainable AI (XAI) has advanced numerous post‐hoc and inherently interpretable techniques ~\cite{meng2023locatingeditingfactualassociations}, such as feature attribution (e.g., SHAP, LIME) and self‐explaining models, to clarify which inputs influence outputs. However, these approaches stop at the model boundary and do not illuminate the causal mechanisms within complex architectures. Mechanistic Interpretability (MI) addresses this gap by treating internal components like layers, attention heads, and neurons as discrete causal units, probed through interventional and observational methods to reveal how information flows and decisions emerge.

Existing XAI work in TSC seldom moves beyond surface-level attributions~\cite{islam2024interpretingtimeseriestransformer}.
Moreover, despite significant progress in MI for NLP transformers, Time Series Transformer models (TSTs) pose distinct challenges: the absence of discrete tokens, the fluid nature of temporal dependencies, and variable sequence lengths complicate direct application of language‐based techniques. This raises the central question of our work:

\begin{quote}
\textit{What novel insights can Mechanistic Interpretability methods, specifically, activation patching, attention saliency, and sparse autoencoders, yield when adapted to Time Series Transformers?}
\end{quote}

\noindent To answer this question, we \textbf{contribute} the following\footnote{Source code and extended experimental results are available at our public GitHub repository \cite{kalnare2025tstmi}.}:

\begin{itemize}
  \item We adapt \textit{activation patching}, \textit{attention saliency}, and \textit{sparse autoencoders} to continuous, variable-length TSTs, enabling causal probing at the layer, head, and timestep levels.
  \item We build \textit{directed causal graphs} that stitch these probes together, mapping information flow from input timesteps through attention heads to the class logits.
  \item We demonstrate that a single early-layer patch can recover up to \textbf{0.89} true-class probability in a misclassified instance, exposing discrete, manipulable circuits inside the model.
  \item We provide complementary sparse-autoencoder visualisations that surface class-specific temporal motifs, enriching the causal story with human-readable patterns.
\end{itemize}

Mechanistic Interpretability at this granularity is critical in safety-critical settings, such as ECG triage (the automated prioritization and classification of electrocardiogram signals to detect urgent cardiac arrhythmias), industrial fault detection, and algorithmic trading, because it converts opaque failures into actionable explanations that auditors can trust.

The remainder of the paper is structured as follows. Section~\ref{sec:related_work} discusses background and related work. Section~\ref{sec:methodology} details the proposed methodology, describing the transformer architecture and interpretability techniques. Section~\ref{sec:experimental_setup} outlines the experimental setup, including datasets and metrics. Section~\ref{sec:results} presents our experimental results, followed by a discussion in Section~\ref{sec:discussion}. Finally, Section~\ref{sec:conclusion} concludes with key findings and future directions.

\section{Background and Related Work}
\label{sec:related_work}
This work lies at the intersection of four research threads: time series classification, Transformer architectures, explainable AI, and particularly Mechanistic Interpretability.

Traditional approaches to Time Series Classification (TSC) have relied on distance-based methods (e.g., $k$-Nearest Neighbors with Dynamic Time Warping) and feature-based classifiers such as Catch22~\cite{lubba2019catch22canonicaltimeseriescharacteristics} and ROCKET~\cite{Dempster_2020}. While computationally efficient, these methods depend heavily on handcrafted features and struggle to capture complex temporal dependencies. Deep learning models, particularly Convolutional Neural Networks (CNNs) and Recurrent Neural Networks (RNNs), have addressed some of these limitations but still face challenges in modeling long-range correlations and irregular sampling.

More recently, Transformer-based architectures have been adapted for TSC due to their self-attention mechanism’s ability to model non-local interactions in sequence data~\cite{wen2023transformerstimeseriessurvey,zervaes2021tst}. Empirical studies report that TSTs consistently outperform CNNs and RNNs on multivariate benchmarks, achieving state-of-the-art accuracy with competitive efficiency \cite{wen2023transformerstimeseriessurvey}. Nevertheless, their superior performance comes at the cost of interpretability: the high-dimensional, distributed representations learned by attention layers and feed-forward blocks obscure how temporal features contribute to final decisions.

XAI has traditionally focused on input–output attribution, employing methods such as SHAP~\cite{lundberg2017unifiedapproachinterpretingmodel} and LIME~\cite{ribeiro2016whyitrustyou} to assign importance scores to input features. In time series contexts, these techniques highlight which timesteps or channels influence predictions but do not reveal how internal computations interact to produce outcomes. Observational approaches like attention saliency further inspect self-attention weights, offering insight into routing patterns but lacking causal guarantees~\cite{serrano2019attentioninterpretable}.

MI emerges as a complementary paradigm that probes model internals as causal systems. By treating layers, attention heads, or neurons as discrete intervention targets, MI methods such as activation patching and causal tracing allow researchers to test \textit{necessity} and \textit{sufficiency} of components via denoising and noising interventions~\cite{heimersheim2024useinterpretactivationpatching}. Sparse autoencoders extend this viewpoint by disentangling latent activations into semantically meaningful codes under sparsity constraints~\cite{cunningham2023sparseautoencodershighlyinterpretable}, enabling targeted analysis and potential manipulation of high-level features.

Notably, this mechanistic approach is inspired by neuroscientific interventions and observations. For example, activation patching parallels transcranial magnetic stimulation (TMS) in neuroscience, by both selectively perturbing internal components to reveal causal effects, while observational saliency analyses are conceptually akin to EEG/MEG recordings that map normal and abnormal signal flows in the brain~\cite{akram2015megattention,romero2011brainmapping}. These analogies underscore the value of causal probing alongside attribution for truly understanding complex systems, whether biological or artificial.

\section{Methodology}
\label{sec:methodology}

To effectively apply activation patching to our TST model, the first critical step is to carefully select source and target instances. We define \textbf{clean instances} as those from the test set where the model predicts the correct class with very high confidence ($P(y_{\text{true}}) > 0.95$). Conversely, \textbf{corrupt instances} are those misclassified or predicted with low confidence ($P(y_{\text{true}}) < 0.50$). This criterion ensures a meaningful contrast while maintaining inherent similarities, as both types originate from the same ground-truth class. Consequently, any observed differences primarily reflect the model’s decision-making behavior rather than inherent class-level variance, making the results of subsequent interventions more interpretable.

To quantify the causal effect of interventions precisely, we use the change in the true-class prediction probability:
\begin{equation}
  \Delta P \;=\; P_{\mathrm{patched}}(y_\text{true}) \;-\; P_{\mathrm{orig}}(y_\text{true})\,,
\end{equation}
where $P_{\mathrm{orig}}$ is the baseline probability on the corrupt instance, $P_{\mathrm{patched}}$ is the probability after an interventional patch, and $y_\text{true}$ is the true class of the instance.

\subsection{Activation Patching}
Activation patching is an interventional method in which activations from a clean instance are injected into a corrupt instance at specific model components. We implement this via PyTorch forward hooks, following a “denoising” protocol:
\begin{enumerate}
  \item Run a forward pass on the clean instance, caching activations at the target component (layer, head, or head–timestep).
  \item During the forward pass of the corrupt instance, replace the corresponding activations with cached values.
  \item Compute $\Delta P$ to measure causal influence.
\end{enumerate}
Notably, this technique can utilize logical structures such as \textit{AND} logic (where multiple activations need simultaneous patching) or \textit{OR} logic (where patching individual activations independently restores model accuracy). We apply this procedure at three levels of granularity:
\begin{itemize}
  \item \textbf{Layer‐level}: replace all attention‐head activations in a given encoder layer.
  \item \textbf{Head‐level}: replace activations of a single attention head.
  \item \textbf{Position‐level}: replace the per‐timestep output of a specific head.
\end{itemize}

We systematically identify critical patches by exhaustively evaluating all layer, head, and position combinations. A patch is deemed \textit{critical} when the resulting change $\Delta P$ surpasses a predefined threshold significantly favoring the correct classification. These identified critical patches form the basis for constructing detailed causal graphs, providing insight into the internal decision-making pathways of the model.

\subsection{Attention Saliency}
To complement activation patching, we utilize attention saliency as an observational interpretability tool. Attention saliency clarifies which timesteps each attention head focuses on during prediction. Specifically, we extract raw attention weights $A^{(h)} \in \mathbb{R}^{T\times T}$ for each head $h$ and average them across all query positions:
\begin{equation}
 S_t^{(h)}=\frac{1}{T}\sum_{i=1}^{T}A_{i,t}^{(h)},
\end{equation}
where $S^{(h)}_{t}$ represents the average attention score for each timestep $t$. While attention saliency does not imply causation, it highlights potential candidate timesteps for targeted causal experiments.

\subsection{Sparse Autoencoders for Latent Feature Analysis}
Direct neuron-level analysis in transformer encoder blocks is challenging due to high dimensionality and semantic entanglement among neuron activations. To overcome this, we introduce Sparse Autoencoders (SAEs) trained on internal activations from the first encoder block’s MLP outputs. The SAE consists of an encoder network, converting high-dimensional activations into a sparse code, and a decoder network, reconstructing the original activations from this sparse code. SAE training uses a mean squared reconstruction loss combined with an $L_1$ sparsity penalty:
\begin{equation}
\mathcal{L}_{\mathrm{SAE}} = ||x - \hat{x}||^2_2 + \lambda\sum_j |z_j|,
\end{equation}

\noindent where $x\in\mathbb{R}^{d}$ is the original MLP activation vector at a single timestep, 
$\hat{x}\in\mathbb{R}^{d}$ is its reconstruction produced by the SAE decoder, 
$z\in\mathbb{R}^{H}$ is the sparse code produced by the SAE encoder, 
$j\in\{1,\dots,H\}$ indexes SAE units, 
$\lambda>0$ weights the $\ell_{1}$ sparsity penalty, 
and $\|\cdot\|_{2}$ denotes the Euclidean norm. The sparsity term ensures a compact set of active neurons, aiding disentanglement and interpretability. After training, we analyze top-activating instances and timesteps per SAE neuron, forming hypotheses about semantically meaningful features in the model’s internal representations.

\section{Experimental Setup}
\label{sec:experimental_setup}
Our classifier \(f_\theta\) operates on inputs \(X\in\mathbb{R}^{T\times C}\), where \(T=25\) is the sequence length and \(C=12\) is the number of channels. We use $d$ to denote the model embedding (hidden) dimension used by the convolutional front end, positional embeddings, attention blocks, and MLPs;
$L$ the number of encoder layers; $H$ the number of attention heads per layer; and $K$ the number of target classes. The model consists of:
\begin{enumerate}
  \item \textbf{Convolutional front end:} Three 1D convolutional layers with kernel sizes \(5,3,3\) and hidden‐dimensional transitions \(C\to d/4\), \(d/4\to d/2\), \(d/2\to d\), each followed by batch‐normalization and ReLU.
  \item \textbf{Positional embedding:} A learnable embedding \(P\in\mathbb{R}^{T\times d}\) added to the conv‐net output.
  \item \textbf{Transformer encoder:} \(L=3\) layers of multi‐head self‐attention (\(H=8\) heads) and two‐layer MLP blocks, each with residual connections, dropout, and layer‐normalization.
    The Transformer’s decoder portion is excluded, since its typical role in generating sequential outputs (e.g.\ next‐word prediction in NLP or time‐series forecasting) is unsuitable for fixed‐length classification tasks.  Omitting the decoder roughly halves the total number of parameters, significantly reducing computational cost without sacrificing classification performance.
  \item \textbf{Classification head:} A temporal max‐pool over \(T\) to produce a \(d\)–dimensional vector, followed by a linear layer \(\mathbb{R}^d\to\mathbb{R}^K\) and softmax to yield class probabilities.
\end{enumerate}

We evaluate on the JapaneseVowels benchmark~\cite{kudo1999japanesevowels}, which comprises LPC‐derived 12‐dimensional frames from nine speakers pronouncing Japanese vowels.  To ensure uniform input shape, all sequences are truncated or zero‐padded to \(T=25\) frames.  The dataset splits into 270 training and 370 test samples across \(Y=9\) classes. Table~\ref{tab:dataset} summarizes its key statistics.

\begin{table}[H]
    \centering
      \caption{JapaneseVowels dataset summary.}
      \label{tab:dataset}
    \begin{tabular}{|l | c| c| c| c| c| c| }
        \hline
       \textbf{Dataset} & \textbf{Train }& \textbf{Test} & \textbf{Series Length} &\textbf{ \# Classes} & \textbf{Dimensionality }& \textbf{Type} \\
        \hline
        JapaneseVowels & 270 & 370 & 25 & 9 & 12 & Audio \\
        \hline
    \end{tabular}
\end{table}

We train \(f_\theta\) for 100 epochs using cross‐entropy loss, with a batch size of 4.  Optimization is performed with the Rectified Adam (RAdam) optimizer (learning rate \(10^{-3}\), weight decay \(10^{-4}\)).  These hyperparameters, particularly the small batch size, intensive weight decay, and RAdam, were chosen to ensure stable convergence given the small training set and to promote effective generalization while mitigating overfitting.  

Models are implemented in PyTorch and trained on a single GPU.  We report standard test accuracy for classification performance, and use the true‐class probability change \(\Delta P\) (see Section~\ref{sec:methodology}) as our primary metric for all interpretability experiments.

\section{Results}
\label{sec:results}

\subsubsection*{\textbf{Baseline Performance \& Instance Selection}}\mbox{}\\
Our first objective is to confirm that the trained TST model achieves strong classification performance, ensuring that subsequent interpretability analyses probe meaningful representations rather than noise.

\begin{figure}[ht]
  \centering  \includegraphics[width=0.8\textwidth]{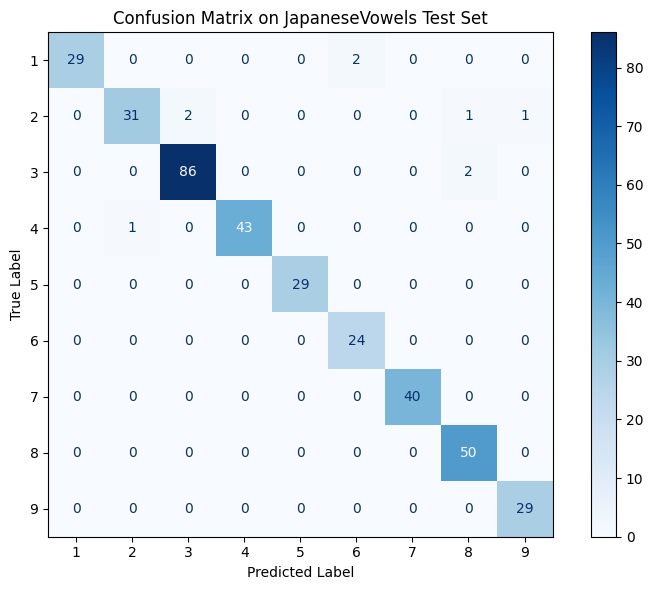}
  \caption{Confusion matrix of the trained TST on the JapaneseVowels test set. Rows are true classes; columns are predicted classes.}
  \label{fig:confusion_matrix}
\end{figure}

Figure~\ref{fig:confusion_matrix} shows the model achieves \textbf{97.57\%} accuracy, with errors confined to a few off-diagonal cells.  Such strong performance ensures that any causal effects we measure reflect learned structure rather than noise.  We therefore select a \emph{clean} instance (ID 33, Class 2, confidence 100.00\%) and a \emph{corrupt} instance (ID 44, Class 2 misclassified as 3 with true class confidence of 10.72\%) for our denoising‐style patching pipeline (Table~\ref{tab:instance_selection}).

\begin{table*}[!htbp]
  \centering
  \caption{Chosen instance pair for activation patching.}
  \label{tab:instance_selection}
  \begin{tabular}{|l|c|c|c|c|}
    \hline
     \textbf{ID} & \textbf{Type}    & \textbf{True Class} & \textbf{Pred. Class} & \textbf{$P(\text{true})$} \\
    \hline
     33 & Clean   & 2          & 2           & 100.00\%           \\
     44 & Corrupt & 2          & 3           & 10.72\%           \\
    \hline
  \end{tabular}
\end{table*}

The pair is ideal for denoising-style activation patching as both belong to the same true class, and the internal activations are aligned in semantic space. The chosen instances' raw time series are visualized in Figure~\ref{fig:sampleinstances} in Appendix \ref{appendixA}. We report results on our primary clean–corrupt pair (IDs 33/44) and replicate every experiment from Sections 5.1–5.5 on a second pair for robustness (see Appendix \ref{appendixActivation}).

\subsubsection*{\textbf{Layer-Level Causal Influence}}\mbox{}\\
We first ask: \emph{Which encoder layer, if patched from the clean to the corrupt instance, most restores correct classification?} To answer this, we replace all attention‐head activations in each layer of the corrupt instance with those from the clean instance, and measure the change in true‐class probability \(\Delta P\).

\begin{figure*}[ht]
  \centering
  \includegraphics[width=1\textwidth]{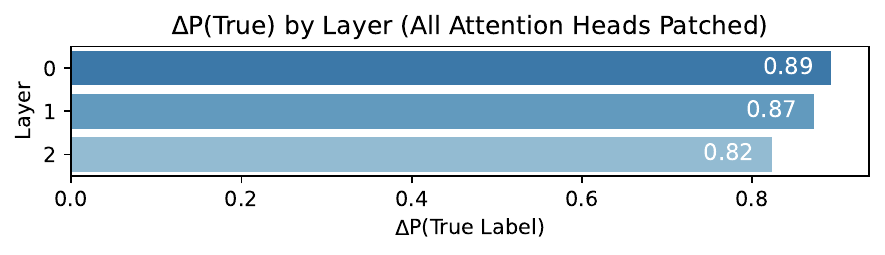}
  \caption{Change in true‐class probability (\(\Delta P\)) when patching all heads of each encoder layer.}
  \label{fig:layer_patching}
\end{figure*}

Figure~\ref{fig:layer_patching} reveals that patching \textbf{Layer 0} yields the largest increase (\(\Delta P\approx0.89\)), with diminishing effects in Layers 1 and 2. This indicates that the earliest attention block carries the most critical causal signal, motivating a finer‐grained head‐level analysis.

The gradual decline in \(\Delta P\) across deeper layers indicates a \emph{distributed} representation: all layers encode useful information, but the causal signal is strongest at the model’s entry point. This observation motivates our next step, drilling down to the head‐level within Layer 0 to pinpoint the individual components driving the bulk of this effect.  

\subsubsection*{\textbf{Head-Level Causal Influence}}\mbox{}\\
To refine our causal analysis, we next inspect the influence of each individual attention head. Rather than patching an entire layer, we perform a sweep across all heads \(h\) in each layer: we inject its activations from the clean instance into the corrupt instance and measure the resulting true‐class probability \(P(y_{\rm true})\).

\begin{figure}[ht]
  \centering
  \includegraphics[width=1\textwidth]{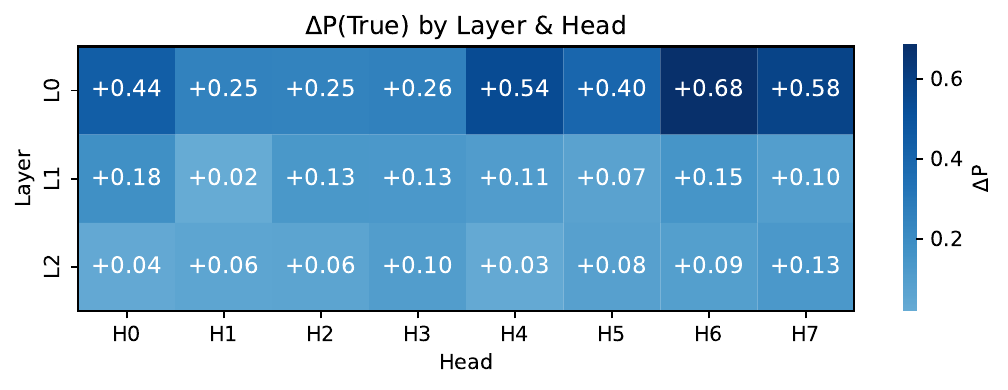}
  \caption{Head‐level activation patching: true‐class probability change \(\Delta P\) after patching each individual attention head.}
  \label{fig:head_patching}
\end{figure}

The corrupt instance’s baseline confidence is \(P(y_{\rm true})=0.1072\). Figure~\ref{fig:head_patching} presents a heatmap where each cell \((\ell,h)\) indicates the model’s true‐class probability change after patching head \(h\) in layer \(\ell\).\footnote{Layers and heads are zero‐indexed; “Layer 0” denotes the first encoder block.}

A close examination of Figure~\ref{fig:head_patching} reveals that causal influence is heavily concentrated in the first encoder layer: patching Heads 0, 4, 5, 6, and 7 individually raises true‐class confidence to above 0.78. In particular, Head 6 boosts \(P(y_{\rm true})\) from 0.1072 to approximately 0.7872 (\(\Delta P\approx0.68\)), echoing the strong Layer 0 effect observed earlier. In contrast, heads in Layers 1 and 2 each yield smaller gains (typically \(\Delta P<0.20\)), with a more uniform distribution of influence and slightly weaker impact in Layer 2 than Layer 1, reinforcing the pattern of diminishing causal strength in deeper blocks. These results point to a sparse internal circuit, where only a handful of heads drive the bulk of the model’s causal computation, guiding us to focus on these key heads for finer positional analyses.


\subsubsection*{\textbf{Position-Level Causal Influence}}\mbox{}\\
\label{sec:position_level}
Having pinpointed Head 6 in Layer 0 as the most influential, we now investigate its \emph{timestep-specific} contributions: \emph{which exact input positions does this head attend to in a causally meaningful way?} To answer this, we patch the per‐timestep output of Head 6 at each position \(t\) (from 0 to \(T-1\)) independently into the corrupt instance and record the resulting change \(\Delta P_t\) in true‐class probability.

\begin{figure}[ht]
  \centering
  \includegraphics[width=1\textwidth]{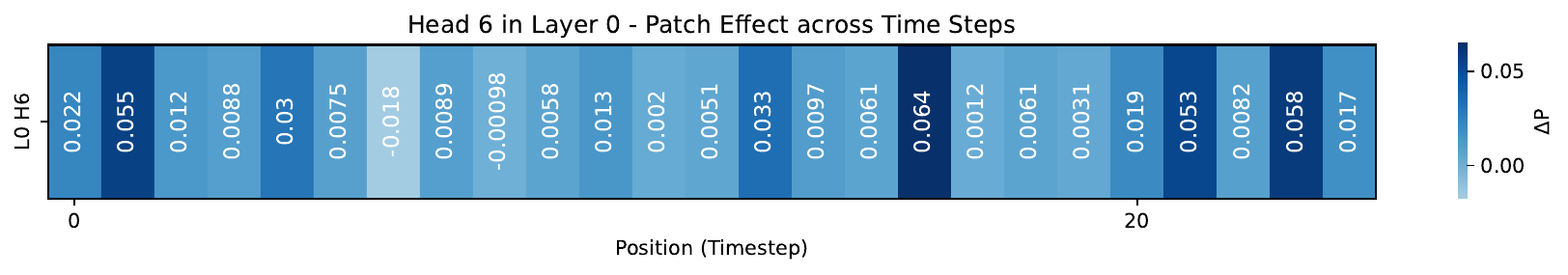}
  \caption{True‐class probability \(\Delta P_t\) after patching each timestep of Layer 0 Head 6.}
  \label{fig:position_patching}
\end{figure}

Figure~\ref{fig:position_patching} reveals a highly non‐uniform distribution of causal impact across timesteps. Approximately half of the positions yield negligible effects (\(\Delta P_t < 0.01\)), and five additional positions barely exceed this threshold. Intriguingly, two positions (6 and 8) produce slightly negative \(\Delta P_t\), indicating that in isolation they introduce interference rather than denoising. In contrast, four positions: 1, 16, 21, and 23, stand out, each contributing \(\Delta P_t > 0.05\). These peaks highlight the “when” of Head 6’s causal intervention: specific segments of the input sequence are critical for steering the model back toward the correct classification.

Importantly, summing these individual contributions 
\[
\sum_{t=0}^{T-1}\Delta P_t \approx 0.43
\]
underestimates the full‐head effect (\(\Delta P_{\text{full head}} \approx 0.68\)). This discrepancy underscores that Head 6’s overall influence is \emph{not} simply additive across timesteps. Instead, it arises from nonlinear synergies enabled by self‐attention, residual connections, and layer normalization, components that integrate information across positions. Consequently, while individual patches reveal candidate “hotspots”, only the simultaneous reinstatement of the entire head’s activation fully recovers the lost confidence, as multiple positions interact constructively (or destructively) when combined. 


\subsubsection*{\textbf{Attention Saliency}}\mbox{}\\
To complement our interventional patching analyses, we apply an observational saliency method to the same head. Specifically, we extract raw self‐attention weights \(A^{(h)}_{i,t}\) from Head 6 in Layer 0 and average over all query positions \(i\) to compute a timestep saliency score \(S^{(h)}_t\). We then overlay these scores on the raw input sequences of the clean and corrupt instances to visualize where the head attends.

\begin{figure}[ht]
  \centering
  \includegraphics[width=1\textwidth]{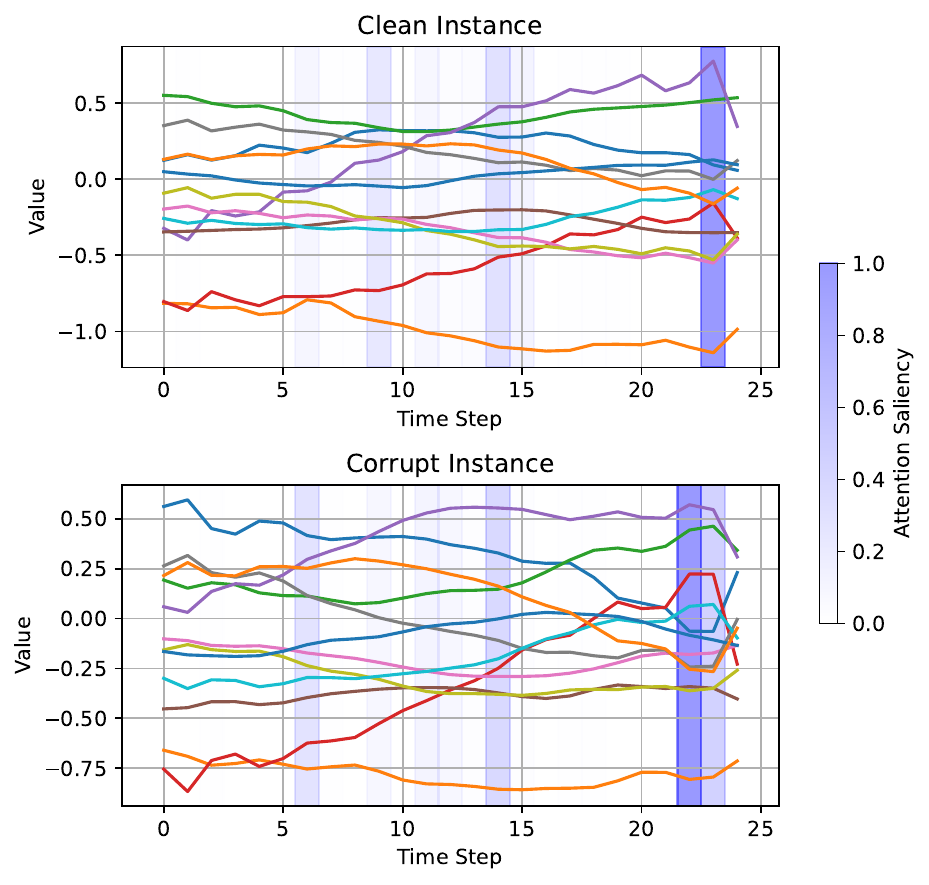}
  \caption{Attention saliency scores of Layer 0 Head 6 overlaid on the raw input series for the clean (top) and corrupt (bottom) instances.}
  \label{fig:attention_saliency}
\end{figure}

Figure~\ref{fig:attention_saliency} shows that saliency is highly concentrated at a few timesteps, most notably some of the final positions. However, it is important to emphasize that attention saliency does \emph{not} imply causation: attention weights can reflect routing or normalization behaviors rather than true feature importance, a limitation well documented in prior NLP studies \cite{jain2019attentionexplanation}. Nonetheless, saliency remains a useful diagnostic, efficiently highlighting candidate positions for more targeted mechanistic interventions.

\subsubsection*{\textbf{Accumulating Top-$k$ Critical Patches}}\mbox{}\\
To evaluate how influential components interact when combined, we conduct a controlled multi-patching experiment. Starting from the most influential (layer, head, position) triplets, which were identified via individual $\Delta P$ scores, we sequentially apply patches from the clean instance into the corrupt instance.

For each $k \in \{1,\dots,10\}$, we apply the top-$k$ critical patches and observe the resulting change in predicted probability for the true class. Table~\ref{tab:multi_patch_deltas} summarizes the results.

\begin{table}[ht]
    \centering
        \caption{Results of applying top-k critical patches. Shows cumulative $\Delta P$ and final predicted probability for the true class as more patches are applied. \label{tab:multi_patch_deltas}}
    
        \begin{tabular}{| c|c|c| }
            \hline
            \textbf{Top-k Patches} & $\Delta P$(True) & $P$(True) \textbf{Final} \\
            \hline
            1  & 0.1270 & 0.2342 \\
            2  & 0.1716 & 0.2788 \\
            3  & 0.2058 & 0.3131 \\
            4  & 0.3055 & 0.4128 \\
            5  & 0.4324 & 0.5396 \\
            \textbf{6}  & \textbf{0.3712} & \textbf{0.4784} \\ 
            7  & 0.4502 & 0.5575 \\
            8  & 0.4835 & 0.5908 \\
            9  & 0.5772 & 0.6844 \\
            10 & 0.6375 & 0.7448 \\
            \hline
        \end{tabular}

\end{table}

The results show that the model's confidence increases steadily as more top-ranked patches are applied, but not monotonically. Notably, applying the 6th patch results in a \textbf{drop in confidence} compared to the 5-patch version, despite being ranked as a top contributor when evaluated in isolation.

This effect illustrates a key point already discussed previously: when multiple internal components are activated simultaneously, their influence may interfere, overlap, or even cancel out due to the model’s nonlinear architecture. Components that are helpful alone may become redundant or conflicting when patched in tandem with others.

\subsubsection*{\textbf{Causal Graph Construction}}\mbox{}\\
To synthesize our interventional and observational findings into a unified representation, we build a directed \emph{causal graph} that traces the flow of influence from input timesteps through internal attention heads to the final class prediction. We organize nodes into three tiers: (i) \emph{Input nodes} \(\mathrm{T}_t\), \(t\in\{0,\dots,T-1\}\), for each timestep in the time‐series instance (green), (ii) \emph{Internal nodes} \(\mathrm{L\ell H h}\) for each attention head \(h\) in layer \(\ell\) (blue) and (iii) \emph{Output nodes} \(\mathrm{C}_y\) for each class \(y\) (orange). Each patch corresponds to two directed edges: a timestep‐to‐head edge weighted by \(\Delta P_{t\to(\ell,h)}\) and a head‐to‐class edge weighted by \(\Delta P_{(\ell,h)\to\mathrm{class}}\).

\begin{figure}[htbp]
  \centering
  \includegraphics[width=1\textwidth]{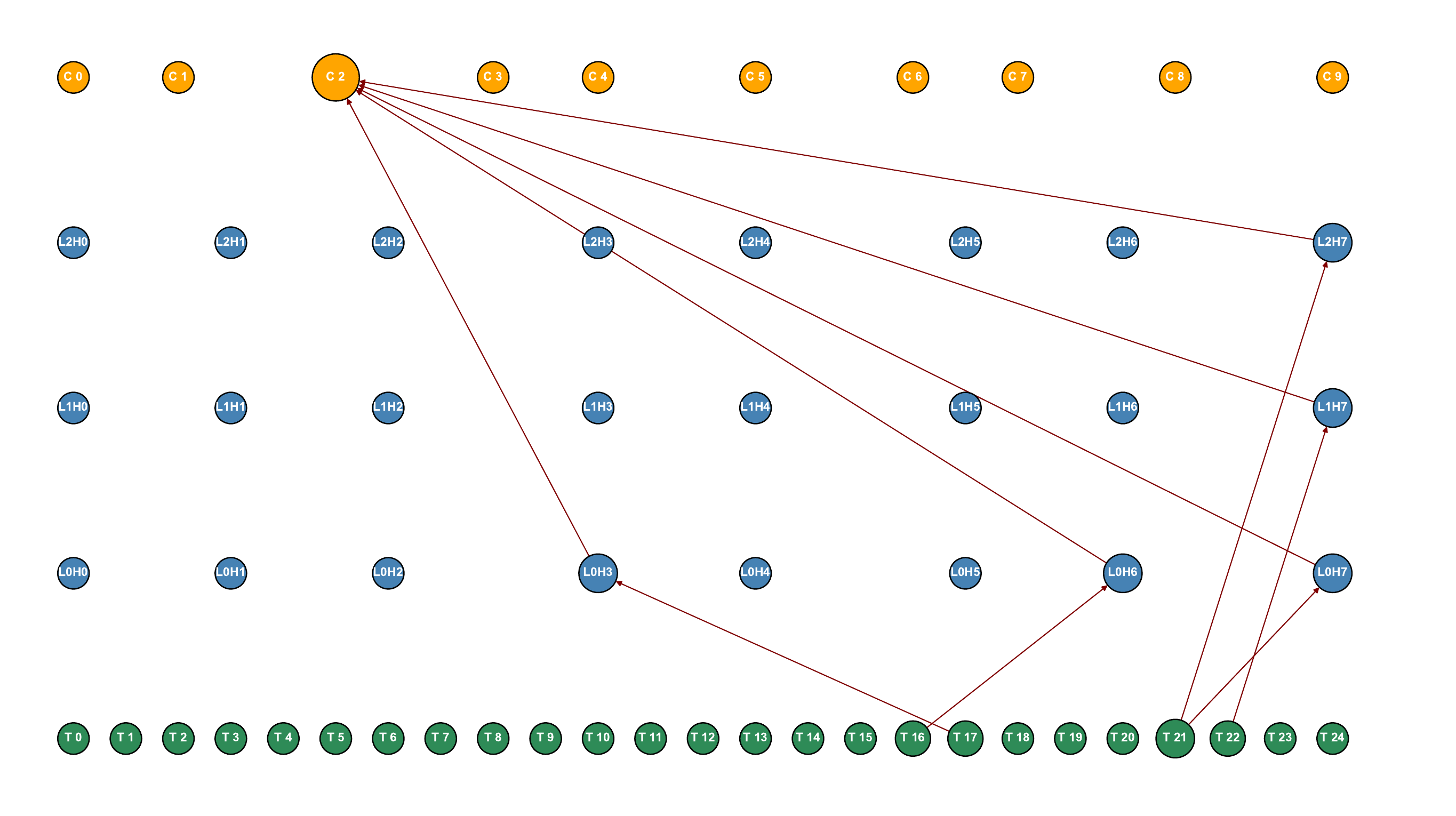}
  \caption{Causal graph of the top-5 most influential patches (timesteps \(\to\) heads \(\to\) class).}
  \label{fig:causal_graph_top5}
\end{figure}

Figure~\ref{fig:causal_graph_top5} shows the \textbf{top-5 patches}, those with the largest individual \(\Delta P\) values, forming a minimal causal circuit that restores \(P(y_{\rm true})\) to 0.5396 (above 0.5).

Notably, three of these patches involve Head 7 in Layers 0, 1, and 2, while the other two involve Heads 6 and 3 in Layer 0. Table~\ref{tab:top5patches} lists these patches in descending order of \(\Delta P\).

\begin{table}[htbp]
  \centering
  \caption{Top-5 critical patches and their \(\Delta P\) contributions.}
  \label{tab:top5patches}
  \begin{tabular}{|c|c|c|}
    \hline
    \textbf{Timestep} & \textbf{Head (L\(\ell\)H\(h\))} & \(\Delta P\) \\
    \hline
    T 21 & L0H7 & 0.1270 \\
    T 21 & L1H7 & 0.0870 \\
    T 22 & L2H7 & 0.0810 \\
    T 16 & L0H6 & 0.0640 \\
    T 17 & L0H3 & 0.0620 \\
    \hline
  \end{tabular}
\end{table}

\begin{figure}[htbp]
  \centering
  \includegraphics[width=1\textwidth]{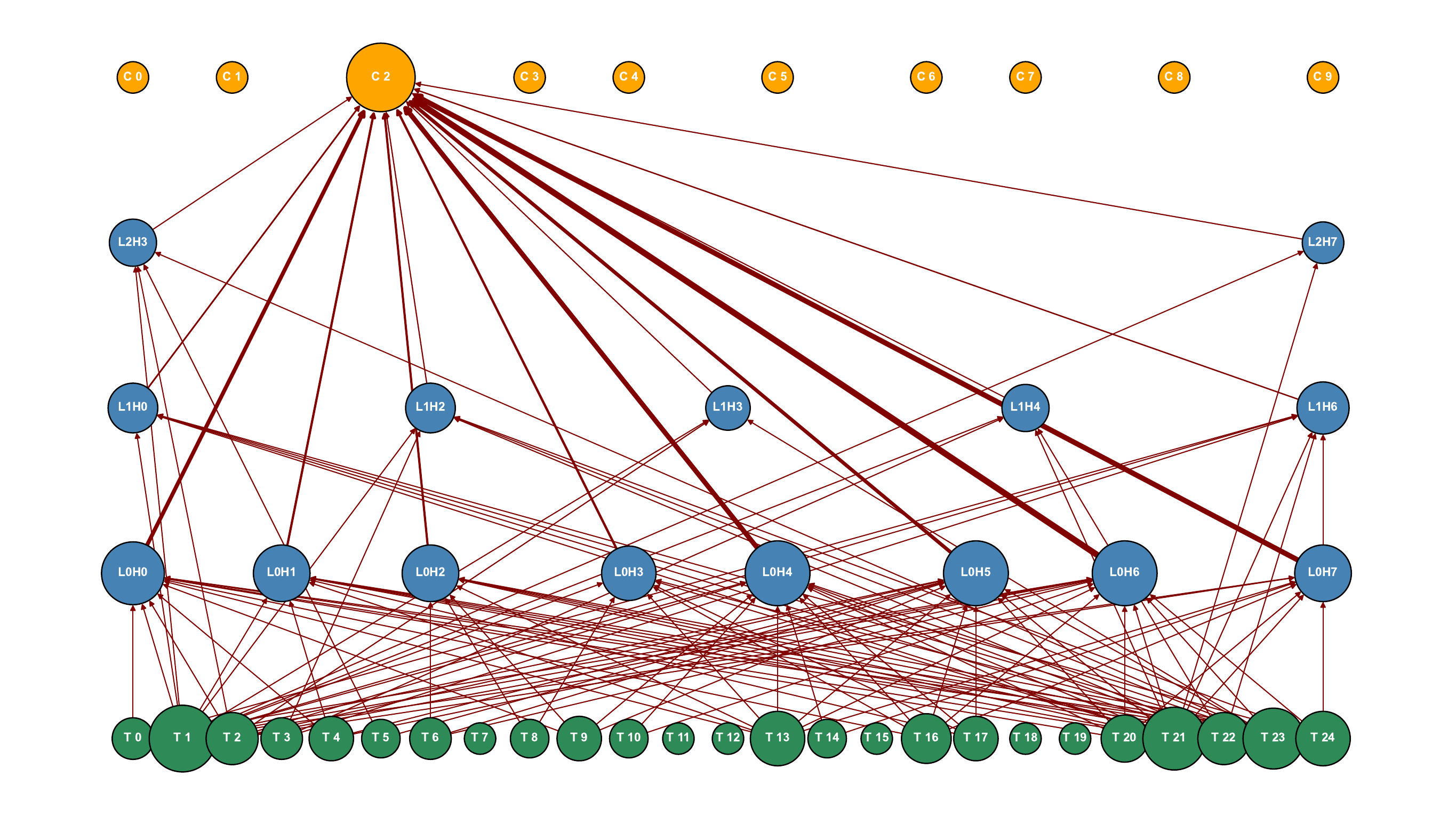}
  \caption{Thresholded causal graph: head \(\to\) class edges with \(\Delta P\ge0.10\), timestep \(\to\) head edges with \(\Delta P\ge0.01\).}
  \label{fig:causal_graph_threshold}
\end{figure}

While this sparse view highlights the most pivotal pathways, it omits moderate but meaningful contributions. To capture a broader circuit, we construct a \textbf{thresholded graph} by including all head‐to‐class edges with \(\Delta P\ge0.10\) and all timestep‐to‐head edges with \(\Delta P\ge0.01\), retaining only heads that satisfy the former condition. Figure~\ref{fig:causal_graph_threshold} shows this denser graph, revealing additional connections, particularly from timesteps near the end of the sequence, into key heads.

\begin{figure}[htbp]
  \centering
  \includegraphics[width=1\textwidth]{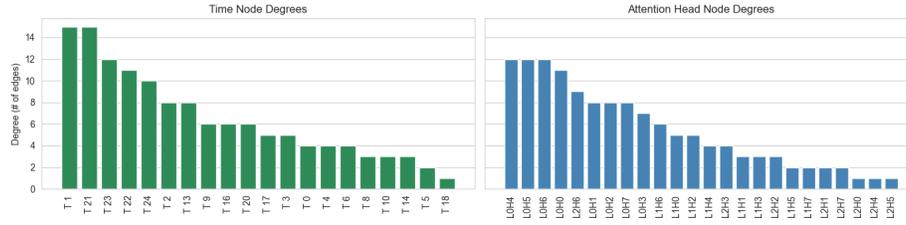}
  \caption{Degree centrality in the thresholded causal graph. \emph{Left:} outgoing degree of each timestep. \emph{Right:} incoming degree of each head.}
  \label{fig:node_degree}
\end{figure}

To quantify the prominence of each node in the thresholded graph, we compute degree centrality: the number of outgoing edges per timestep and the number of incoming edges per head (excluding head‐to‐class edges). Figure~\ref{fig:node_degree} plots these degrees. 

On the input side, late timesteps (21–24) dominate outgoing connections, with an additional early peak at Time 1, indicating non‐uniform reliance on sequence positions. On the internal side, eight of the top‐ten most connected heads reside in Layer 0, reinforcing our earlier observations that the first encoder layer concentrates causal influence.

\subsection{Provisional Sparse Autoencoder Insights}
\label{sec:sae_insights}

To uncover whether the transformer’s internal activations contain disentangled, semantically meaningful features, we train a SAE on the output of the MLP block (\texttt{linear2}) in the first encoder layer.  The SAE encoder maps each activation vector \(x\in\mathbb{R}^d\) at each timestep into a sparse code \(z\in\mathbb{R}^H\), where \(H\) is the number of SAE neurons.  A strong \(L_1\) penalty on \(z\) encourages each code to activate only a few neurons, facilitating interpretability.

\begin{figure*}[ht]
  \centering
\includegraphics[width=1\textwidth]{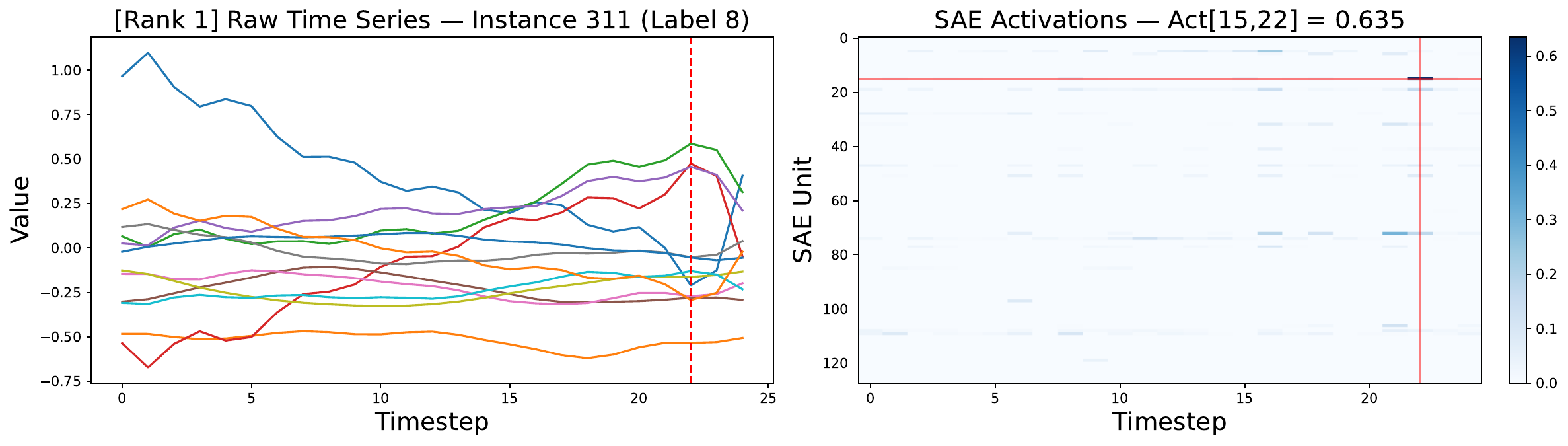}
  \caption{Top‐activating test instances for SAE neuron 15 at timestep 22.  Each overlay shows the raw input time series, with neuron 15’s activation highlighted.  All top‐activating instances belong to Class 8.}
  \label{fig:sae_neuron_example}
\end{figure*}

Figure~\ref{fig:sae_neuron_example} illustrates a representative top‐activating instance for SAE neuron 15 peaking at timestep 22 (the remaining top‐activating instances are shown in Appendix~Figure~\ref{fig:Neuron15}). All of these top‐ranked instances share the ground‐truth label (Class 8), and inspection of their raw inputs reveals a strikingly consistent waveform motif: a pronounced simultaneous peak in Channels 2, 5, and 9 accompanied by a dip in Channel 7 around that timestep. This recurring pattern across multiple speakers suggests that neuron 15 has specialized to detect this class‐specific temporal feature, rather than idiosyncratic noise in any single instance.

\begin{figure*}[ht]
  \centering
  \begin{subfigure}[b]{0.68\textwidth}
    \includegraphics[width=\textwidth]{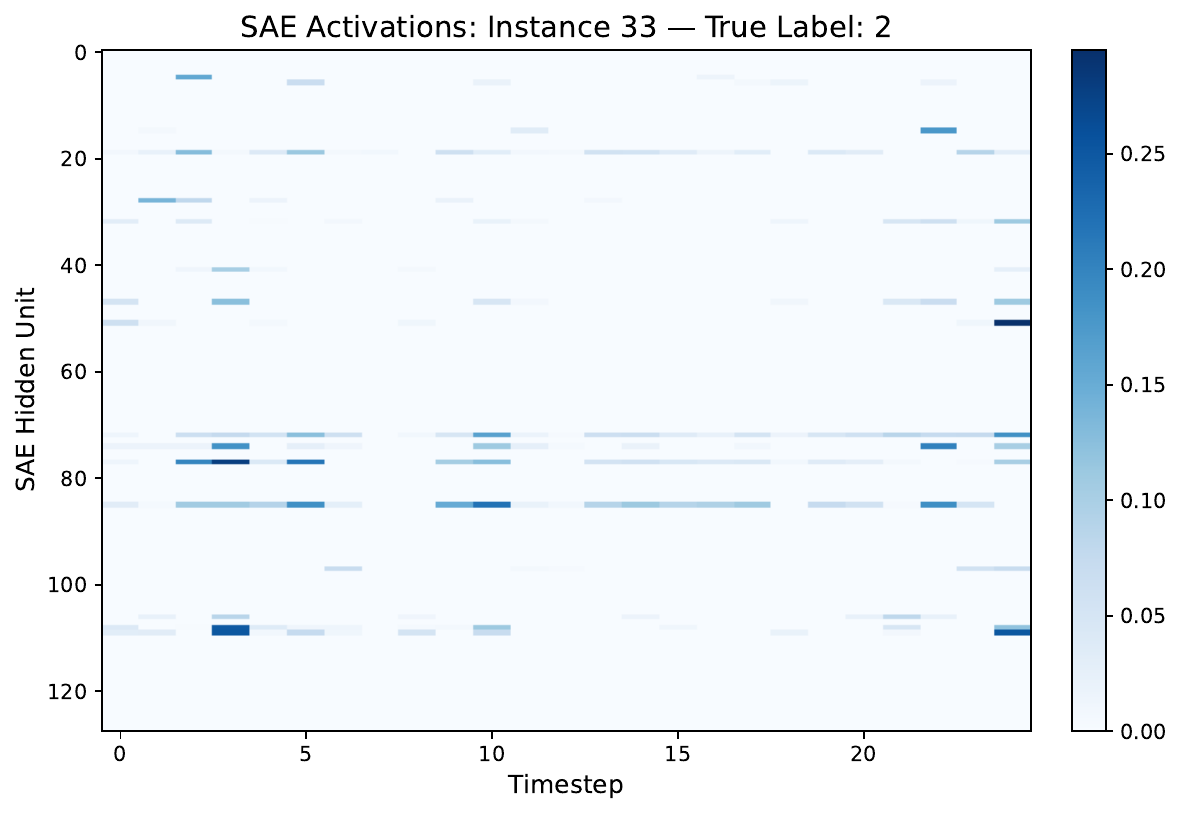}
    \label{fig:sae33}
  \end{subfigure}
  \hfill
  \begin{subfigure}[b]{0.68\textwidth}
    \includegraphics[width=\textwidth]{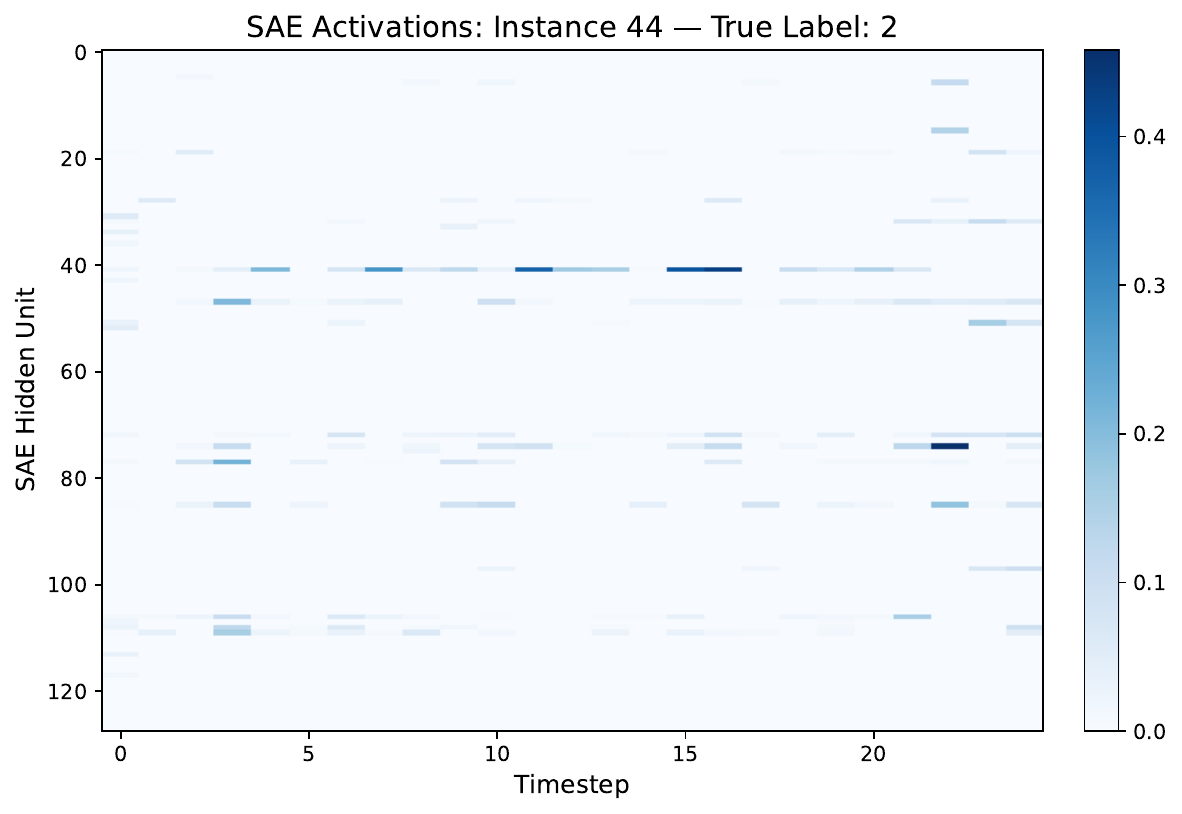}
    \label{fig:sae44}
  \end{subfigure}
  \caption{SAE activation heatmaps (neurons \(\times\) timesteps) for the clean and corrupt instances.  Darker cells denote higher activation values.}
  \label{fig:sae_comparison}
\end{figure*}

To see how this sparse code behavior contrasts with our clean–corrupt pair, we plot full SAE activation heatmaps (neurons \(\times\) timesteps) for Instance 33 (clean, Class 2) and Instance 44 (corrupt, misclassified) in Figure~\ref{fig:sae_comparison}. 

Specifically, Figure~\ref{fig:sae_comparison} shows that in the clean instance several neurons (notably 78 and 85) activate sharply at specific timesteps that align with class‐discriminative patterns, whereas the corrupt instance exhibits a different activation profile, activating neuron 41 more broadly but failing to trigger the combination of neurons that uniquely identify Class 2.  This divergence suggests that misclassification may arise when an instance erroneously engages features characteristic of another class. In Appendix~\ref{appendixA} Figure~\ref{fig:2vs3}, we provide further visualizations showing the activation of these neurons across other Class 2 and Class 3 test samples.

Importantly, SAE neurons do not correspond one‐to‐one with the transformer’s original neurons.  Instead, each SAE neuron represents a learned direction in the transformer’s latent space.  To test the causal role of these sparse features, one could encode the transformer activations into \(z\), manually amplify a chosen dimension \(z_j\), decode back to \(\hat{x}\) via the SAE decoder, and patch \(\hat{x}\) into the transformer’s MLP at the corresponding layer and timestep. Thus, measuring the resulting change in \(P(y_{\rm true})\) would reveal the sufficiency of individual sparse features.

\section{Discussion}
\label{sec:discussion}
Mechanistic Interpretability has emerged as a critical component for trustworthy AI, especially in safety‐critical domains such as healthcare and finance where understanding internal decision pathways can build trust and facilitate risk mitigation. Our experiments demonstrate that MI techniques originally developed for NLP transformers: activation patching, attention saliency, and sparse autoencoders, can be effectively adapted to TSTs, yielding fine‐grained insights into which layers, heads, timesteps, and latent features drive model decisions. The resulting causal graphs and sparse feature visualizations (Section~\ref{sec:sae_insights}) illustrate multilevel interpretability in TSTs and highlight both shared patterns across layers and instance‐specific motifs.

Despite these promising results, several limitations constrain our current pipeline. First off, MI methods remain labor‐intensive: selecting representative clean–corrupt pairs, choosing granularity levels, and interpreting interventional outcomes require substantial domain expertise. Second, our patching experiments focus solely on attention heads; other components like MLP blocks, residual streams, positional embeddings, and individual neurons likely encode additional causal signals that remain unexplored.  Third, unlike NLP, where tokens often correspond to semantic concepts, time series features lack clear semantics, making interpretation of causal motifs (e.g., peaks, periodicities, anomalies) inherently more challenging. This raises questions of \emph{universality}, whether the circuits we uncover generalize across datasets or reflect dataset-specific idiosyncrasies. Finally, our multi‐patching results show non‐additive effects; interventions can interfere or synergize due to the transformer’s complex interplay of multi‐head attention, residual connections, and layer normalization, complicating efforts to isolate independent causal factors.

To address these limitations, we envision several avenues for future research. First, expanding patching to include MLP activations, residual streams, and embeddings will provide a more comprehensive view of TST internals. Second, scaling analysis to multiple clean–corrupt pairs across classes and datasets, especially higher-dimensional or irregular real‐world time series, will test the generality of discovered circuits. Third, integrating optimization‐driven search (e.g., evolutionary algorithms) could automatically identify minimal, non‐redundant patch sets, refining causal graph construction.  Fourth, developing higher‐level metrics beyond \(\Delta P\), as well as visualization and summarization techniques (e.g., clustering pathways, generating narrative descriptions), will enhance interpretability for non‐expert stakeholders. Finally, leveraging MI for post‐hoc error correction, dynamically patching critical components at inference time, offers a promising route to improve model robustness without retraining.

By systematically extending and automating these MI frameworks, we can move toward interpretable, reliable TSTs that not only perform well but also reveal their inner workings in a way that aligns with human understanding and ethical standards.

\section{Conclusion}
\label{sec:conclusion}

This work set out to determine whether Mechanistic Interpretability techniques, originally developed for NLP transformers, can be meaningfully adapted to TSTs. Through a series of structured experiments on a speaker‐identification task using the JapaneseVowels dataset, we have shown that activation patching at the layer, head, and timestep levels reveals clear causal effects on model predictions. Patching individual attention heads and even specific timestep outputs can substantially restore correct classification in mispredicted instances, demonstrating that TST internals encode discrete, manipulable circuits akin to those observed in language models.

By synthesizing these interventional findings into causal graphs, we uncovered interpretable pathways from input timesteps through attention heads to class logits, and we observed nonlinear synergies and interference when combining multiple patches. Complementary observational methods, attention saliency, and sparse autoencoders, provided additional insights into where and when the model attends and what latent features it extracts. In particular, sparse autoencoders uncovered class‐selective temporal motifs, highlighting the potential for disentangled feature discovery in time series domains.

While we stop short of a full reverse‐engineering of TSTs, our results establish a robust proof‐of‐concept: MI tools can help in gaining understanding about the hidden logic of deep time series models. Ultimately, embedding Mechanistic Interpretability into TSTs can enhance transparency, accountability, and reliability of AI systems deployed in critical real‐world applications.  

\newpage

\bibliographystyle{splncs04}
\bibliography{references}

\appendix
\section{Causal Influence via Activation Patching}\label{appendixActivation}
To verify that our activation‐patching pipeline generalizes beyond the primary example (Section 5), we repeat the full sequence of analyses on a second clean–corrupt pair. All panels below mirror Sections 5.1–5.5.

\subsection{Pair 2: Instance Selection and Overview}\label{appendix:pair2}

Table \ref{tab:instance_selectionAp} lists the chosen pair. The “clean” instance 123 is classified correctly as class 3 with almost perfect confidence, while the “corrupt” instance 114 (also true label 3) is misclassified as class 8 with only 18.5 \% true‐class probability.

\begin{table}[ht]
  \centering
  \caption{Second clean–and–corrupt instance pair for robustness testing.}
  \label{tab:instance_selectionAp}
  \begin{tabular}{|l|c|c|c|c|}
    \hline
     \textbf{ID} & \textbf{Type}    & \textbf{True Class} & \textbf{Pred. Class} & \textbf{$P(\text{true})$} \\
    \hline
     123 & Clean   & 3          & 3           & 99.99\%           \\
     114 & Corrupt & 3          & 8           & 18.49\%           \\
    \hline
  \end{tabular}
\end{table}

\begin{figure}[ht]
  \centering
  \includegraphics[width=\textwidth]{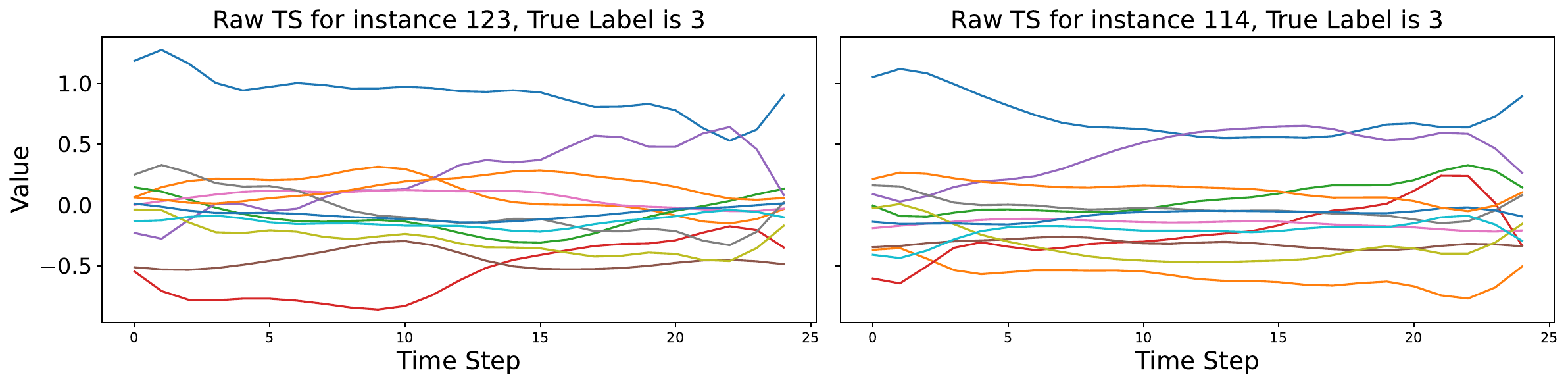}
  \caption{Raw multivariate time series for instances 123 (clean) and 114 (corrupt), 12 channels over 25 timesteps.}
  \label{fig:sampleinstancesap}
\end{figure}

Figure \ref{fig:sampleinstancesap} shows the raw 12‐channel time series for both instances. 

\subsection{Layer‐wise patching.}
\begin{figure}[ht]
  \centering
  \includegraphics[width=\textwidth]{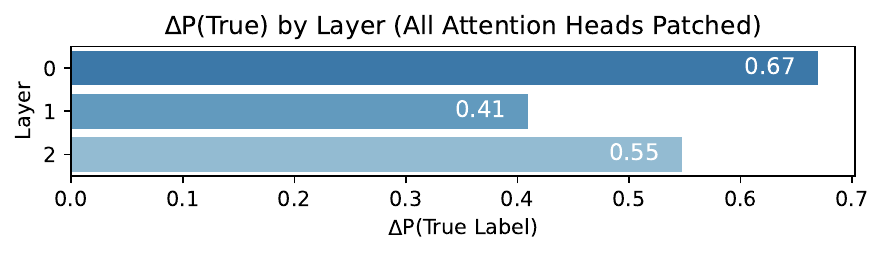}
  \caption{True‐class probability after patching all heads of each encoder layer.}
  \label{fig:layer_patchingap}
\end{figure}

In Figure \ref{fig:layer_patchingap}, we patch all heads in each encoder layer of the clean into the corrupt instance and plot the resulting true‐class probability. Layer 0 again exerts the strongest causal influence (increase from 0.185 to 0.855, \(\Delta P_t\)=0.67), confirming early layers’ dominance.

\subsection{Head‐wise patching.}
\begin{figure}[ht]
  \centering
  \includegraphics[width=\textwidth]{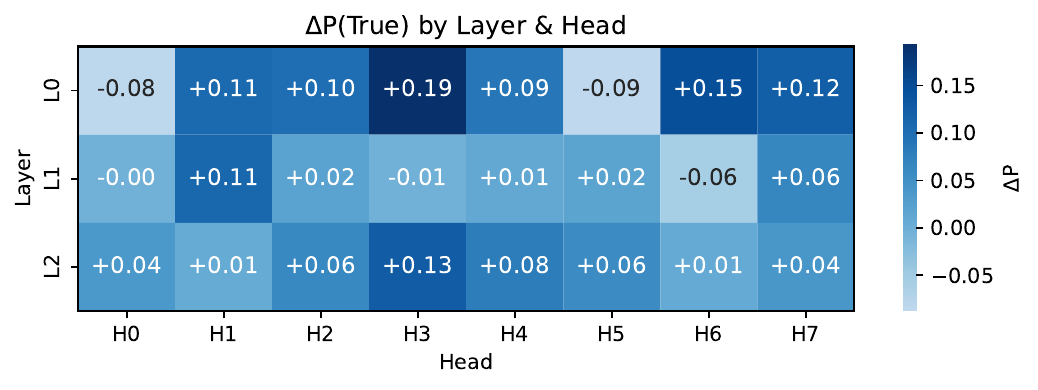}
  \caption{Post‐patching true‐class probability for each attention head.}
  \label{fig:head_patchingap}
\end{figure}

Breaking down Layer 0 into individual heads (Figure \ref{fig:head_patchingap}), head 3 contributes the largest improvement (\(\Delta P_t \approx +0.19\)), followed by heads 1 and 7. Other layers show far smaller or negligible effects.

\subsection{Position‐wise patching.}
\begin{figure}[ht]
  \centering
  \includegraphics[width=\textwidth]{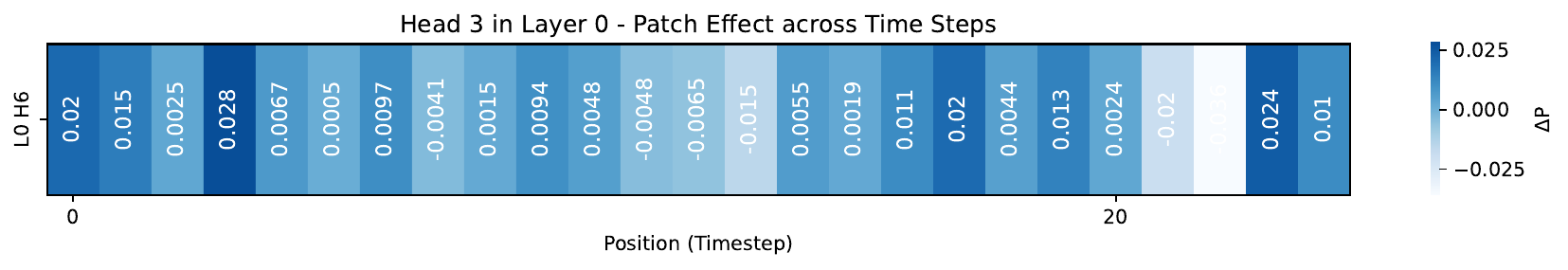}
  \caption{True‐class probability change \(\Delta P_t\) after patching each timestep of Layer 0 Head 3.}
  \label{fig:position_patchingap}
\end{figure}

Focusing on Layer 0 Head 3, Figure \ref{fig:position_patchingap} plots true‐class probability after patching each timestep in isolation. We again observe “hotspots” around timesteps 0, 3, 17, and 23, where a single‐step patch raises $P(y_{\rm true})$ by more than 0.02 (baseline=0.1849).

\subsection{Attention‐saliency comparison.}
\begin{figure}[ht]
  \centering
  \includegraphics[width=\textwidth]{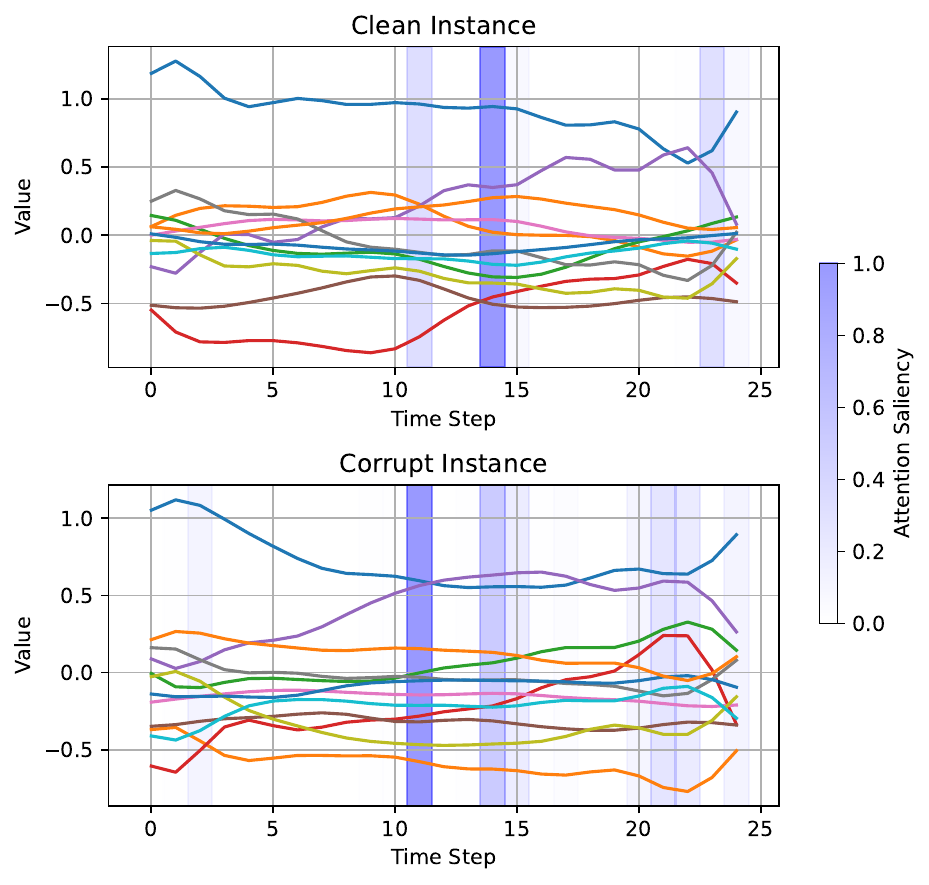}
  \caption{Attention‐saliency scores for Layer 0 Head 3 overlaid on raw input series. Top: clean instance; bottom: corrupt instance.}
  \label{fig:attention_saliencyap}
\end{figure}

Finally, Figure \ref{fig:attention_saliencyap} overlays Layer 0 Head 3’s observational attention‐saliency scores on both inputs. 

\newpage

\begin{figure}[ht]
\section{Additional Visualizations}\label{appendixA}
  \centering  \includegraphics[width=1\textwidth]{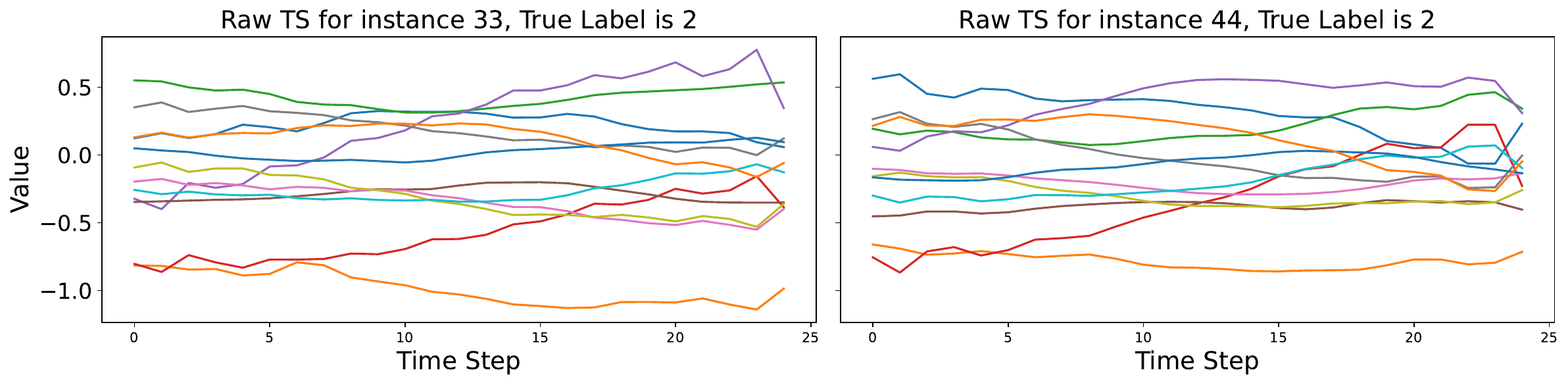}
  \caption{Sample visualizations of multivariate time series instances 33 (clean) and 44 (corrupt) from the JapanseVowels dataset, showing 12-channel signal over 25 time steps.}
\label{fig:sampleinstances}
\end{figure}

\begin{figure}[ht]
  \centering
  \includegraphics[width=0.9\linewidth]{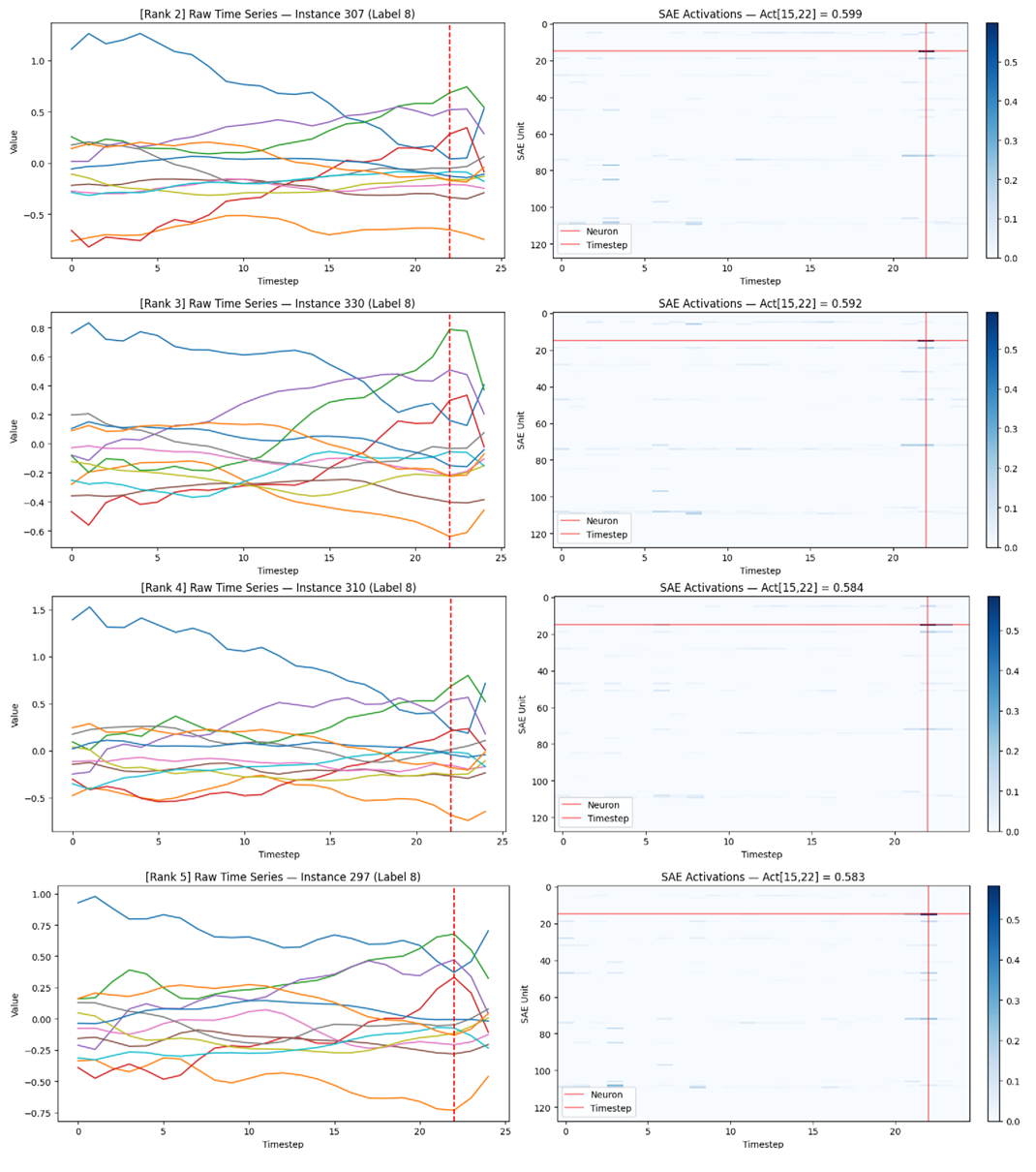}
  \caption{Top-activating test instances for SAE neuron 15 at timestep 22. These instances all belong to Class 8 and exhibit similar temporal patterns, suggesting neuron 15 encodes a class-discriminative feature.}
  \label{fig:Neuron15}
\end{figure}

\begin{figure}[ht]
  \centering
  \includegraphics[width=0.9\linewidth]{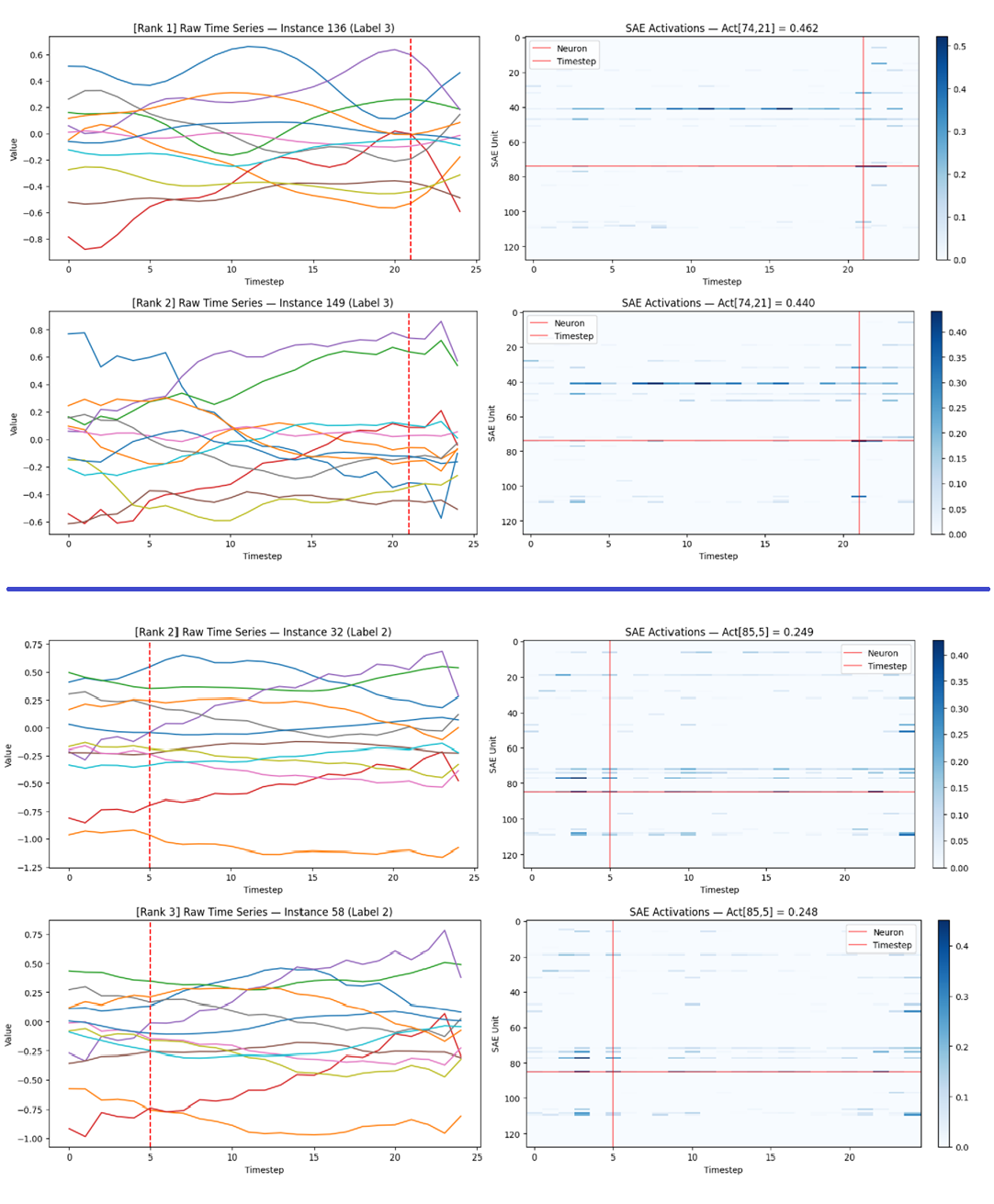}
  \caption{Comparison of sparse activations in Class 2 vs Class 3 instances. Highlights class-selective neurons such as neuron 85 and 78 (Class 2) and neuron 41 (Class 3).}
  \label{fig:2vs3}
\end{figure}

\end{document}